\documentclass{article}
\usepackage{spconf}
\usepackage{amsmath}
\usepackage{graphicx}
\usepackage{xcolor}
\usepackage{amssymb}
\usepackage{tabularray}
\usepackage{subcaption}
\usepackage{fancyhdr}
\usepackage{tikz}
\usetikzlibrary{calc}

\usepackage[acronym]{glossaries}
\newacronym{pnp}{PnP}{plug-and-play}
\newacronym{sisr}{SISR}{single image super-resolution}
\newacronym{awgn}{AWGN}{additive white Gaussian noise}
\newacronym{mse}{MSE}{mean squared error}
\newacronym{psnr}{PSNR}{peak signal-to-noise ratio}
\newacronym{sam}{SAM}{spectral angle mapper}
\newacronym{ssim}{SSIM}{structural similarity index measure}
\newacronym{hqs}{HQS}{half quadratic splitting}
\newacronym{pca}{PCA}{principal component analysis}
\newacronym{hir}{HIR}{hyperspectral image restoration}

\def\x{{\mathbf x}}

\title{Leveraging Pretrained RGB Denoisers for Hyperspectral Image Restoration
}
\name{Daniele Picone, Mohamad Jouni, Mauro Dalla Mura
\thanks{This work is partially supported by grants ANR FuMultiSPOC (ANR-20-ASTR-0006) and ANR  MIAI Cluster (ANR-23-IACL-0006).}
}
\address{Univ. Grenoble Alpes, CNRS, Grenoble INP, GIPSA-Lab, 38000, Grenoble, France}
\fancypagestyle{FirstPage}{
	\fancyhf{}

	\fancyfoot[C]{\footnotesize {© 2026 IEEE. Personal use of this material is permitted.  Permission from IEEE must be obtained for all other uses, in any current or future media, including reprinting/republishing this material for advertising or promotional purposes, creating new collective works, for resale or redistribution to\\servers or lists, or reuse of any copyrighted component of this work in other works.}} 
}

\begin{document}
%

\maketitle
\thispagestyle{FirstPage}
\begin{abstract}
Hyperspectral image restoration faces several challenges, including limited training data, strong sensor specificity, and high spectral dimensionality. These limitations hinder the learning of robust hyperspectral priors, motivating the reuse of priors learned from large-scale RGB data.
In this work, we propose a minimally trained, lightweight adapter that repurposes frozen pretrained RGB denoisers for hyperspectral restoration through a projection mapping.
The method denoises low-dimensional spectral projections and reconstructs the hyperspectral cube through constrained linear aggregation, while preserving plug-and-play compatibility and the stability properties of the underlying RGB denoiser.
Experiments on denoising, deblurring, and super-resolution across multiple datasets demonstrate consistent improvements over hyperspectral-specific baselines, showing the strong transferability of large-scale RGB priors.
\end{abstract}

\section{Introduction}
\label{sec:intro}

Hyperspectral (HS) imaging provides rich spectral information essential for material characterization and quantitative analysis, with applications in remote sensing, medical imaging, and environmental monitoring. However, hyperspectral measurements are inherently sensitive to noise and degradation due to limited photon budgets and sensor-specific acquisition pipelines, making robust image restoration a critical prerequisite.
%

Deep learning has driven significant progress in RGB image restoration, primarily through powerful priors learned from large-scale datasets~\cite{Zhu23, Li25a}. In the hyperspectral domain, a parallel line of research has emerged, focusing on learning spectral-spatial priors tailored to hyperspectral data. These priors have been incorporated into task-agnostic frameworks, including deep image prior-based methods~\cite{Sido19}, \gls{pnp} approaches~\cite{Lai22}, diffusion models~\cite{Pang24, Deng24}, and more recent all-in-one architectures~\cite{Wu25}.
However, hyperspectral restoration methods remain constrained by the scarcity of training data and the need for sensor-specific architectures. As a result, many approaches rely on carefully designed spectral-spatial networks or task-specific models, which often generalize poorly across datasets and restoration tasks.
In contrast, image denoisers developed for RGB images are typically trained on large and diverse datasets and have demonstrated strong generalization properties across a wide range of restoration problems, which yields a simple question.
Can pretrained RGB denoisers be transferred to hyperspectral tasks?
As illustrated in \figurename~\ref{fig:rgb_vs_hs}, a standard RGB denoiser trained on natural images can yield comparable denoising results when applied to selected hyperspectral bands, despite the absence of any hyperspectral training. This suggests that pretrained RGB denoisers encode transferable image priors that may be exploited beyond their original domain, motivating the question of how such models can be systematically leveraged for \gls{hir}.

\def\x{0.48}
\def\y{1.0}
\begin{figure}[t]
		\centering
		\begin{subfigure}{\x\linewidth}
			\centering
			\includegraphics[width=\y\linewidth]{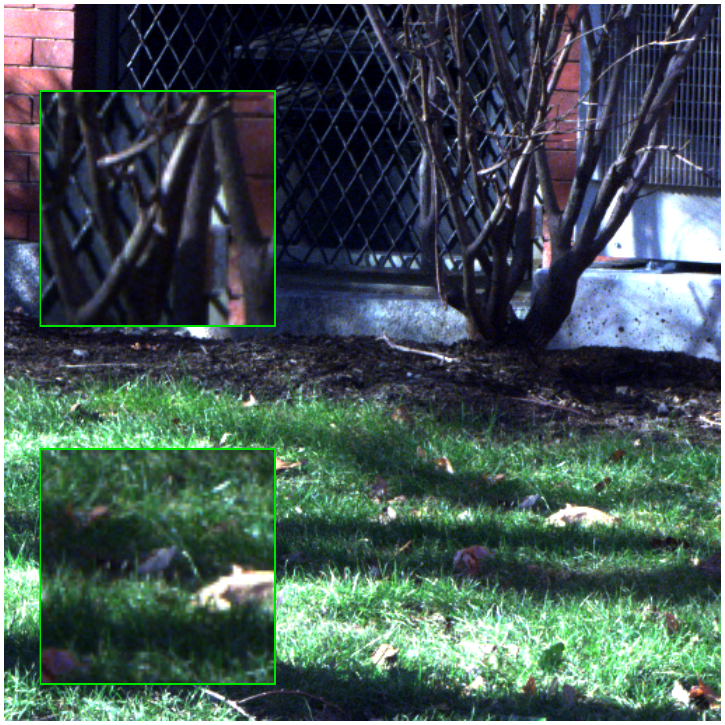}
			\caption{
				\centering
				Reference\newline
				~
			}
		\end{subfigure}\hfill
		\begin{subfigure}{\x\linewidth}
			\centering
			\includegraphics[width=\y\linewidth]{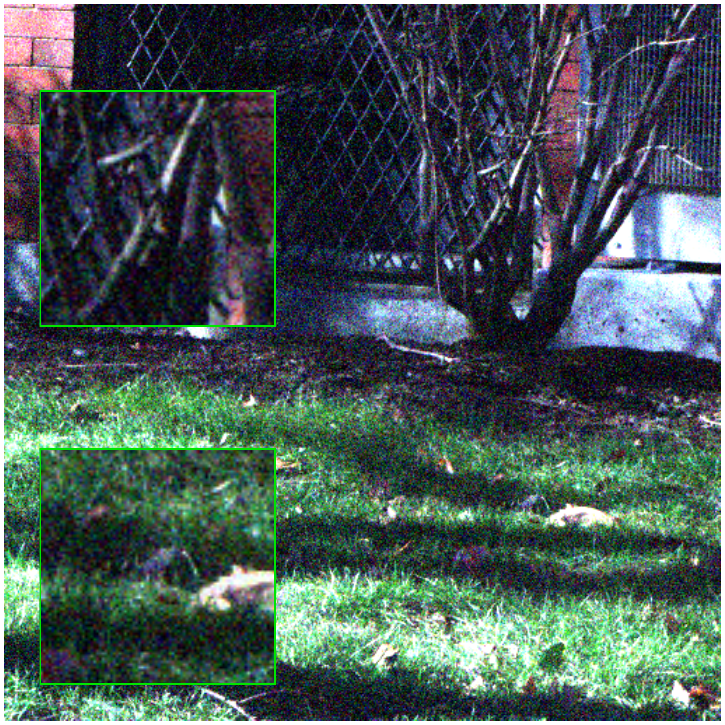}
			\caption{
				\centering 
				Noisy image\newline
				$\mathrm{PSNR} = 20.00 \, \mathrm{dB}$
			}
		\end{subfigure}
		\medskip
		\begin{subfigure}{\x\linewidth}
			\centering
			\includegraphics[width=\y\linewidth]{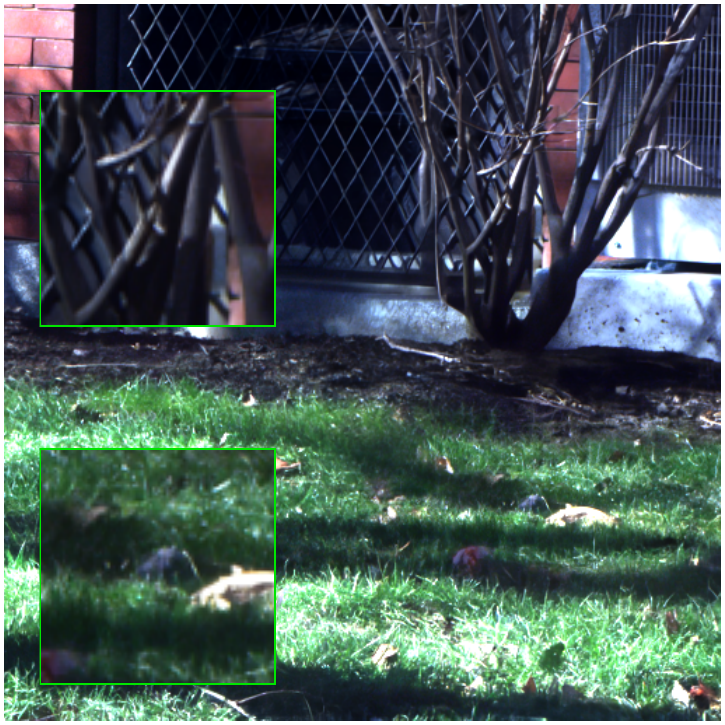}
			\caption{
				\centering 
				RGB denoiser (DRUNet~\cite{Zhan22})\newline
				$\mathrm{PSNR} = 30.42 \, \mathrm{dB}$
			}
		\end{subfigure}\hfill
		\begin{subfigure}{\x\linewidth}
			\centering
			\includegraphics[width=\y\linewidth]{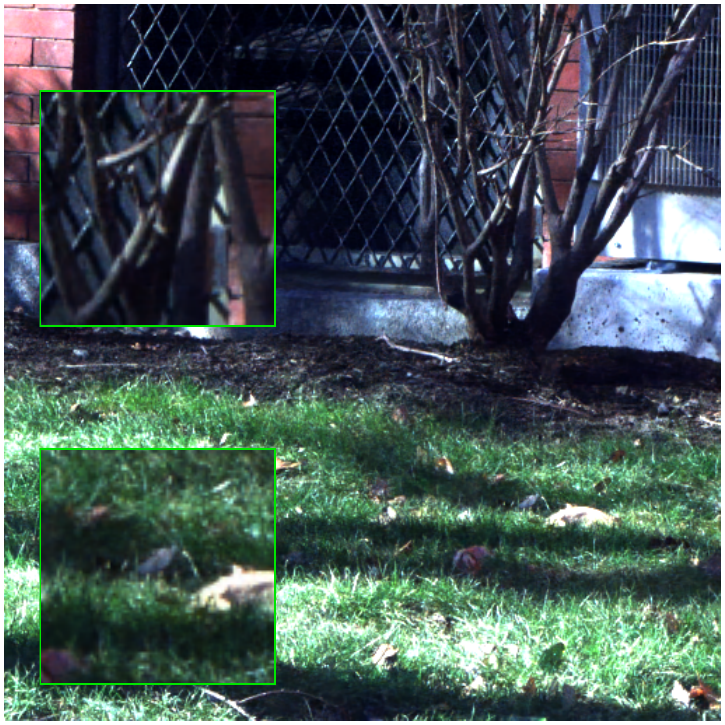}
			\caption{
				\centering 
				HS denoiser (GRUNet~\cite{Roy24})\newline
				$\mathrm{PSNR} = 30.35  \, \mathrm{dB}$
			}
		\end{subfigure}
	\caption{Comparison of pretrained RGB and HS denoising results for an image in the Harvard~\cite{Chak11} dataset with \acrshort{awgn} and standard deviation $\sigma=0.10$. RGB denoising is performed over the bands $\{24,\,14,\,4\}$, while HS denoising processes the full image, with the same bands displayed for visualization. PSNR values reported only for the three selected bands.}
	\label{fig:rgb_vs_hs}
\end{figure}

In this work, we propose a simple and principled framework to reuse pretrained RGB or monochromatic denoisers for hyperspectral restoration.
Inspired by the adapter framework for classification~\cite{Pere22}, our approach wraps a frozen low-dimensional denoiser within a lightweight adapter, that (i) projects hyperspectral data into low-dimensional subspaces compatible with the denoiser, and (ii) aggregates the resulting estimates back into the hyperspectral domain.
The adapter is linear, preserves the stability (e.g. non-expansiveness) of the underlying denoiser, and requires little or no hyperspectral training. The resulting hyperspectral denoiser can seamlessly be integrated into \gls{pnp} optimization schemes, enabling its use across multiple inverse problems. We demonstrate the effectiveness of the proposed approach on denoising, deblurring, and \gls{sisr} tasks, and provide ablation studies that highlight the role of spectral projections and pretrained RGB priors.
Unlike DPHSIR~\cite{Lai22}, which relies on a dedicated hyperspectral denoiser, our method adapts a frozen RGB prior. Unlike RGB-to-HS reconstruction methods~\cite{Akht20} or RGB-assisted diffusion models~\cite{Deng24}, it restores hyperspectral data directly without training a joint hyperspectral architecture. To the best of our knowledge, fully frozen RGB denoisers agnostic to hyperspectral data have not been explicitly explored for hyperspectral restoration.
The main contributions of this work are: (i) a projection-based adapter framework that lifts pretrained RGB or monochromatic denoisers to  \gls{hir}; (ii) a constrained linear design that preserves stability and \gls{pnp} compatibility.

\section{Proposed method}

We aim to reconstruct an ideal hyperspectral image $\mathbf{x}\in\mathcal{X}\subseteq\mathbb{R}^{N \times M \times C}$, where $N$ and $M$ are the spatial dimension and $C$ is the number of spectral bands.
We consider a widely used degradation model, where the observation $\mathbf{y}\in\mathcal{Y}$ is given by:
\begin{equation}
	\mathbf{y} = \mathbf{A}\mathbf{x} + \mathbf{n}
	\label{eq:noise}
\end{equation}
where $\mathbf{A}:\mathcal{X}\rightarrow \mathcal{Y}$ is a linear operator modeling the degradation of the latent image, and $\mathbf{n}\in\mathcal{Y}$
denotes a noise realization drawn from a stochastic distribution.

Many classical inverse problems can be formulated within this framework, where 
$\mathbf{A}$ may represent an identity, blurring, or spatial downsampling. The corresponding reconstruction tasks are commonly referred to as denoising, deblurring and \gls{sisr}, respectively.
We focus on the denoising case ($\mathbf{A}$ is an identity), while other inverse problems are addressed via \gls{pnp}, as shown in the experimental section.
We assume access to a pretrained denoiser $\mathcal{D}^\mathit{l}_\sigma: \mathbb{R}^{M \times N \times c} \rightarrow \mathbb{R}^{M \times N \times c}$ designed for monochromatic ($c=1$) or RGB ($c=3$) images, trained under an \gls{awgn} model with standard deviation $\sigma$ and kept frozen throughout all experiments. Our goal is to leverage this low-dimensional denoiser to construct a hyperspectral denoiser $\mathcal{D}_\sigma^\mathit{h}: \mathbb{R}^{M \times N \times C} \rightarrow \mathbb{R}^{M \times N \times C}$.

\begin{figure}
	\includegraphics[width=\linewidth]{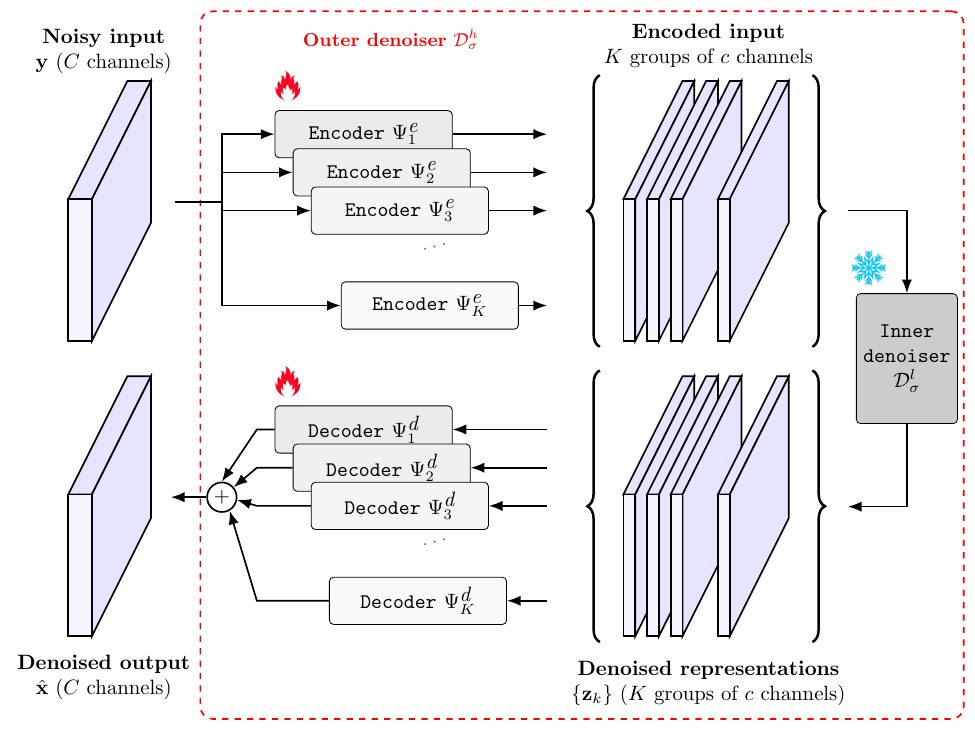}
	\caption{Architecture for hyperspectral adaptation. The overall denoiser is outlined by a red dashed box. A fire and ice symbol respectively denote trainable and frozen layers.}
	\label{fig:architecture}
\end{figure}

In \figurename~\ref{fig:architecture}, we define the hyperspectral denoiser as an aggregation of $K$ projected denoising groups, each operating in a $c$-dimensional latent space compatible with the pretrained denoiser. Formally,
\begin{equation}
	\mathcal{D}_\sigma^{\mathit{h}}(\mathbf{y}) = \sum_{k=1}^{K} \Psi_k^\mathit{d}(\mathcal{D}_\sigma^{\mathit{l}}(\Psi_k^\mathit{e}(\mathbf{y})))
\end{equation}
where $\Psi_k^\mathit{e}: \mathbb{R}^{M \times N \times C} \rightarrow \mathbb{R}^{M \times N \times c}$ and a $\Psi_k^\mathit{d}: \mathbb{R}^{M \times N \times c} \rightarrow \mathbb{R}^{M \times N \times C}$ respectively denote the encoder and decoder associated with the $k$-th projection.
Each pair $(\Psi_k^\mathit{e}, \Psi_k^\mathit{d})$ defines a distinct projection / reconstruction subspace mapping, while the same pretrained denoiser $D_\sigma^\mathit{l}$ is reused across all groups.

In this work, we restrict both the encoder and decoder to be linear operators acting along the spectral dimension. For each $k$, let $\mathbf{E}_k\in\mathbb{R}^{c \times C}$  and $\mathbf{F}_k\in\mathbb{R}^{C \times c}$ denote the encoding and decoding matrices, respectively.
The encoder is applied pixel-wise, as $(\Psi_k^\mathit{e}(\mathbf{y}))_{m,n,:} = \mathbf{E}_k\mathbf{y}_{m,n,:} \in \mathbb{R}^c\,.$
%
The projected images are processed by the pretrained denoiser, producing denoised projections $\mathbf{z}_k$, which are mapped back to the hyperspectral domain using a decoding matrix $\mathbf{F}_k\in\mathbb{R}^{C \times c}$, such that $
	(\Psi_k^\mathit{d}(\mathbf{z}_k))_{m,n,:} = \mathbf{F}_k\mathbf{z}_{m,n,:} \in \mathbb{R}^C\,.
$
For notational convenience, we concatenate the individual encoders and decoders into global matrices:
\begin{align}
	\mathbf{E} &= [\mathbf{E}_1^T, \mathbf{E}_2^T, ..., \mathbf{E}_K^T]^T \in \mathbb{R}^{(Kc) \times C}\\
	\mathbf{F} &= [\mathbf{F}_1, \mathbf{F}_2, ..., \mathbf{F}_K] \in \mathbb{R}^{C \times (Kc)}
\end{align}

Under this notation, the overall encoding produces a stacked latent representation in $\mathbb{R}^{Kc}$, while the decoder aggregates the reconstructed hyperspectral contributions from all projection groups.
The estimation (or design) of $\mathbf{E}$ and $\mathbf{F}$ fully specifies the proposed hyperspectral denoiser.

To ensure a well-posed reconstruction, we impose two fundamental properties that must hold when the inner denoiser $\mathcal{D}_\sigma^l$ is replaced by the identity operator.
Specifically, the global decoder must invert the global encoder, i.e.,
$\mathbf{F}\mathbf{E}= \mathbf{I}_C$
and the singular values of the encoder $\mathbf{E}$ must be bounded to prevent excessive noise amplification. A natural way to enforce both properties is to constrain $\mathbf{E}$ to have orthonormal columns (i.e., $\mathbf{E}^T\mathbf{E}=\mathbf{I}_{C}$) and to define the decoder $\mathbf{F}$ as the Moore-Penrose pseudo-inverse of $\mathbf{E}$.

This condition also preserves the Lipschitz constant of the inner denoiser. If $\mathcal{D}_\sigma^\mathit{l}$ is $L$-Lipschitz and $\|\mathbf{E}\|=1$, then $\mathcal{D}_\sigma^\mathit{h}$ is also $L$-Lipschitz:

\begin{equation}
\begin{aligned}
    \|\mathcal{D}_\sigma^\mathit{h}(\mathbf{y}')-\mathcal{D}_\sigma^\mathit{h}(\mathbf{y})\|
    &\le \|\mathbf{E}^T\|\,
    \|\mathcal{D}_\sigma^\mathit{l}(\mathbf{E}\mathbf{y}')
      -\mathcal{D}_\sigma^\mathit{l}(\mathbf{E}\mathbf{y})\| \\
    &\le L\|\mathbf{E}\mathbf{y}'-\mathbf{E}\mathbf{y}\| \\
    &\le L\|\mathbf{E}\|\,\|\mathbf{y}-\mathbf{y}'\| \\
    &\le L\|\mathbf{y}-\mathbf{y}'\| .
\end{aligned}
\label{eq:liptchitz}
\end{equation}
Thus, the projection-aggregation wrapper does not tighten the Lipschitz requirements usually involved in \gls{pnp} fixed-point convergence analyses~\cite{Ryu19}.
We enforce these properties through a QR parametrization. An unconstrained matrix $\tilde{\mathbf{E}}\in\mathbb{R}^{Kc \times C}$ is factorized as $\tilde{\mathbf{E}}=\mathbf{Q}\mathbf{R}$, and we set $\mathbf{E}=\mathbf{Q}$ and $\mathbf{F}=\mathbf{Q}^T$. This guarantees $\mathbf{E}^T\mathbf{E}=\mathbf{I}_C$ when $Kc\ge C$. On the contrary, if $Kc < C$, exact reconstruction is no longer guaranteed and the encoder–decoder pair yields the minimum-norm least-squares reconstruction.


In practice, the matrix $\tilde{\mathbf{E}}$ is optimized during training, while the encoder $\mathbf{E}$ is obtained at each iteration by projecting onto the set of orthonormal matrices via a QR decomposition.
%
%
The encoder is learned by minimizing an \gls{mse} loss between the denoised reconstruction and the ground-truth hyperspectral image:
\begin{equation}
	\mathcal{L}(\tilde{\mathbf{E}})
	=
	\big\|
	\mathbf{F}\,
	\mathcal{D}_\sigma^l(\mathbf{E}\mathbf{y})
	-
	\mathbf{x}
	\big\|_2^2,
\end{equation}
where the encoder $\mathbf{E}$ and decoder $\mathbf{F}$ are implicitly defined from the unconstrained matrix $\tilde{\mathbf{E}}$ via the QR-based parametrization described above. The optimization is carried out with respect to $\tilde{\mathbf{E}}$ using a standard Adam-based scheme.

\begin{figure*}[t]
	\centering	
    \hspace{0.1pt}
    \raisebox{0.1\height}{
        \includegraphics[width=0.025\linewidth]{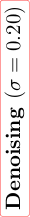}
    }
	\hspace{0.05pt}
	\begin{subfigure}{0.19\linewidth}
		\centering
      		\includegraphics[width=\linewidth]{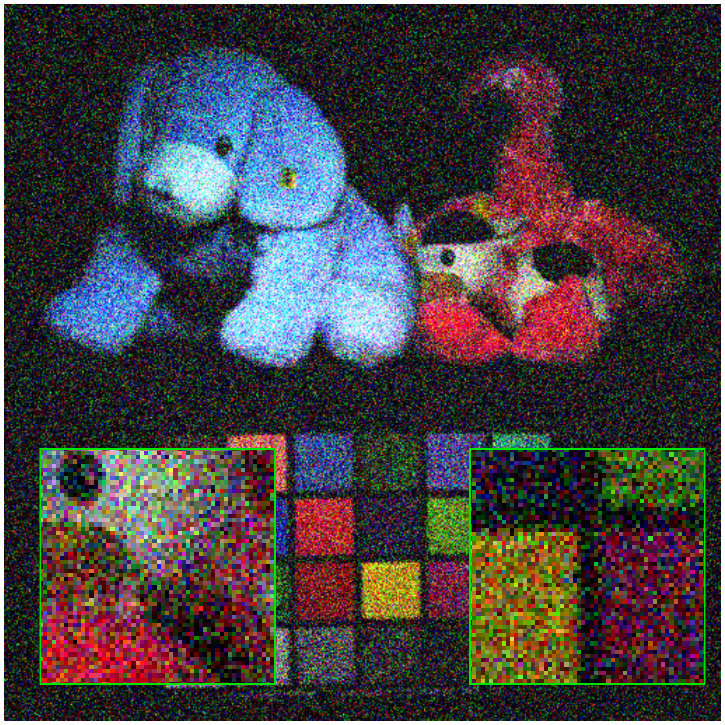}
		\caption{Observation $\mathbf{y}$}
	\end{subfigure}\hfill
	\begin{subfigure}{0.19\linewidth}
		\includegraphics[width=\linewidth]{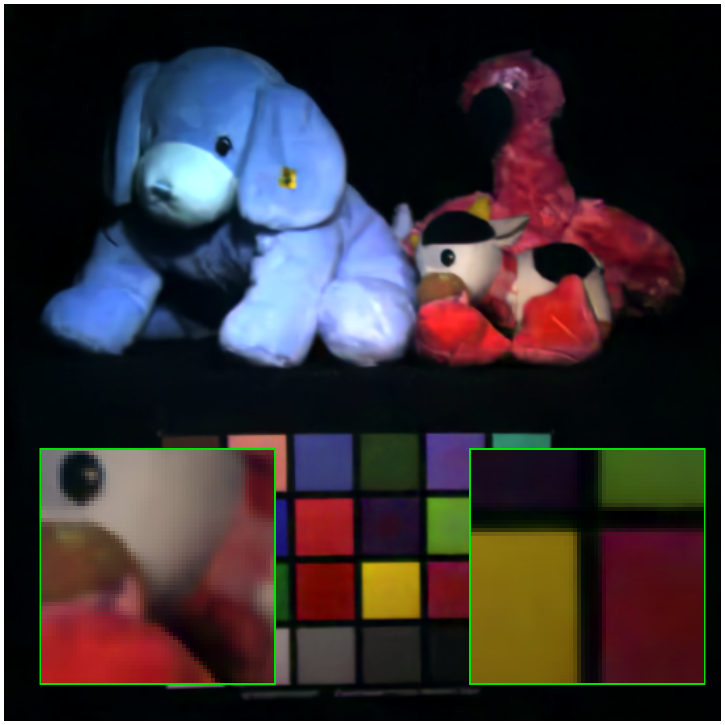}
		\caption{T3SC~\cite{Bodr21}}
	\end{subfigure}\hfill
	\begin{subfigure}{0.19\linewidth}
		\centering
		\includegraphics[width=\linewidth]{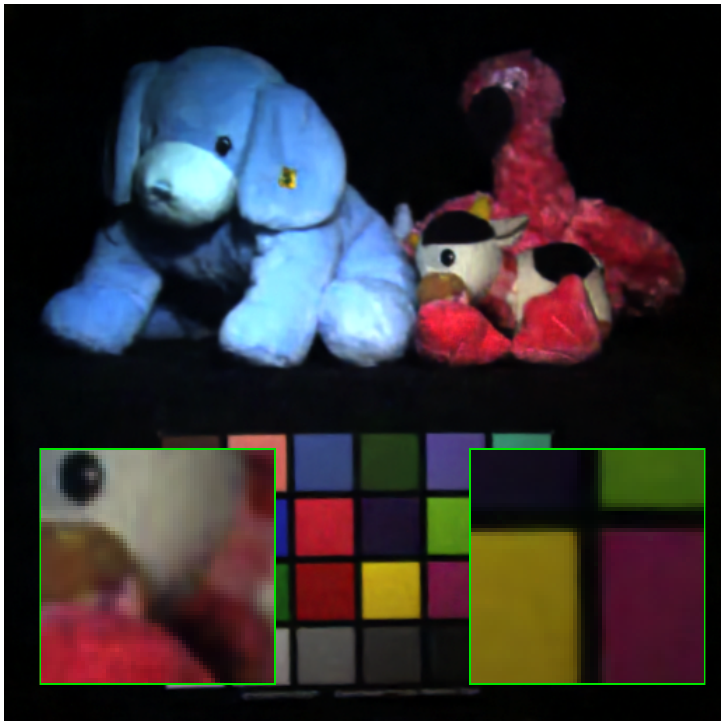}
		\caption{GRUNet~\cite{Roy24}}
	\end{subfigure}\hfill
	\begin{subfigure}{0.19\linewidth}
		\centering
		\includegraphics[width=\linewidth]{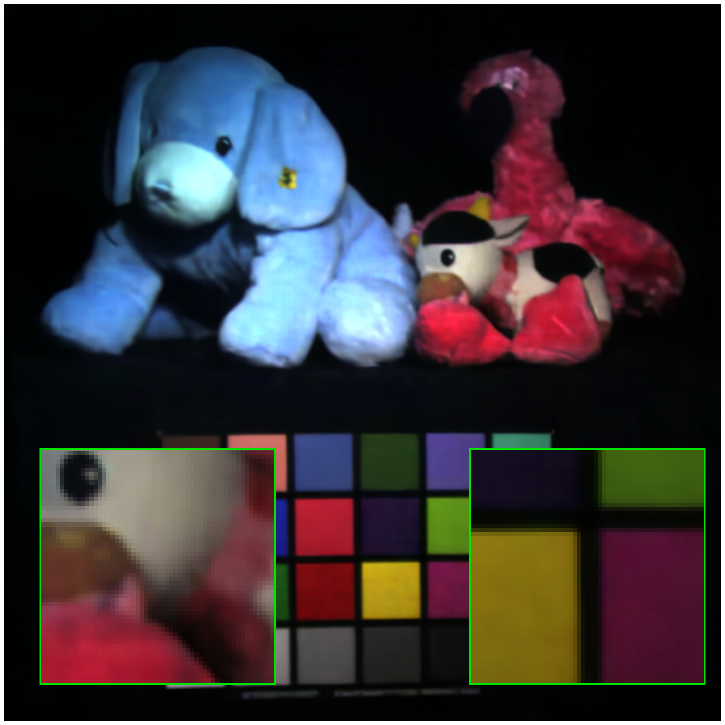}
		\caption{SERT~\cite{Li23d}}
	\end{subfigure}\hfill	
	\begin{subfigure}{0.19\linewidth}
		\centering
		\includegraphics[width=\linewidth]{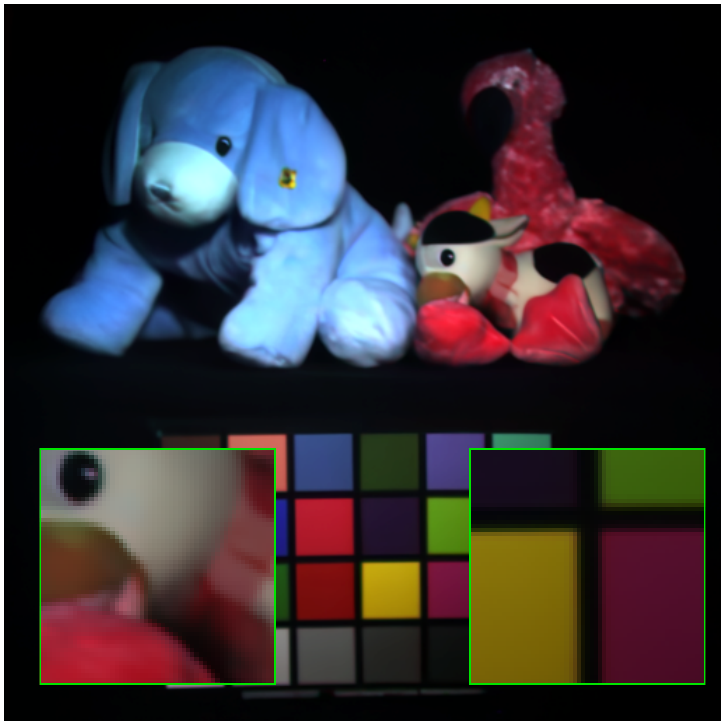}
		\caption{Proposed}
	\end{subfigure}\hfill
	\smallskip
	\hspace{0.1pt}
    \raisebox{0.1\height}{
        \includegraphics[width=0.025\linewidth]{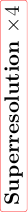}
    }
	\hspace{0.05pt}
	\begin{subfigure}{0.19\linewidth}
		\centering
		\includegraphics[width=\linewidth]{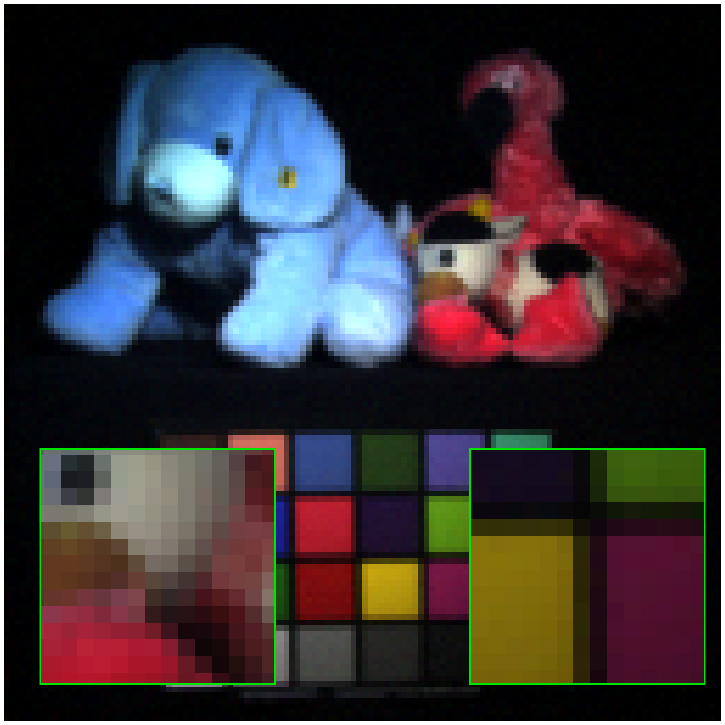}
		\caption{Observation $\mathbf{y}$}
	\end{subfigure}\hfill
	\begin{subfigure}{0.19\linewidth}
		\centering
		\includegraphics[width=\linewidth]{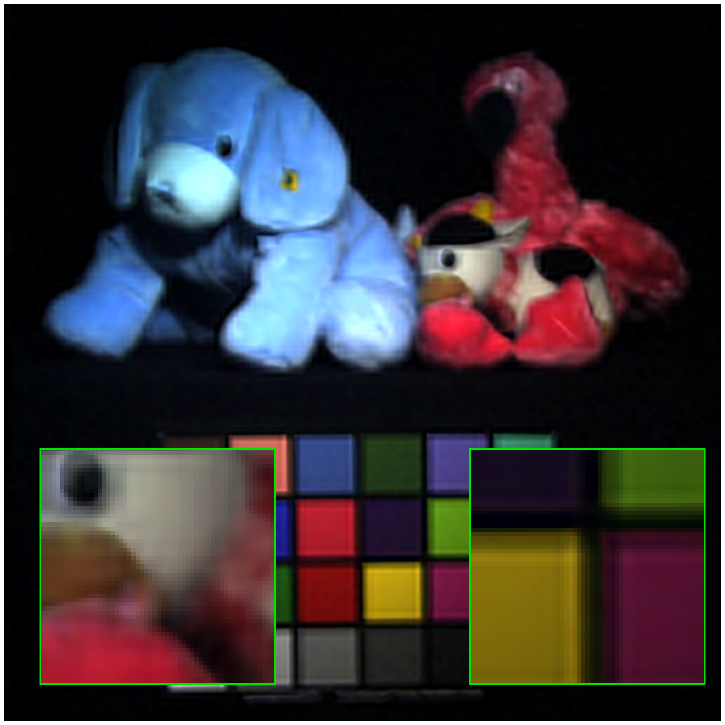}
		\caption{HSISR~\cite{Li22d}}
	\end{subfigure}\hfill
	\begin{subfigure}{0.19\linewidth}
		\centering
		\includegraphics[width=\linewidth]{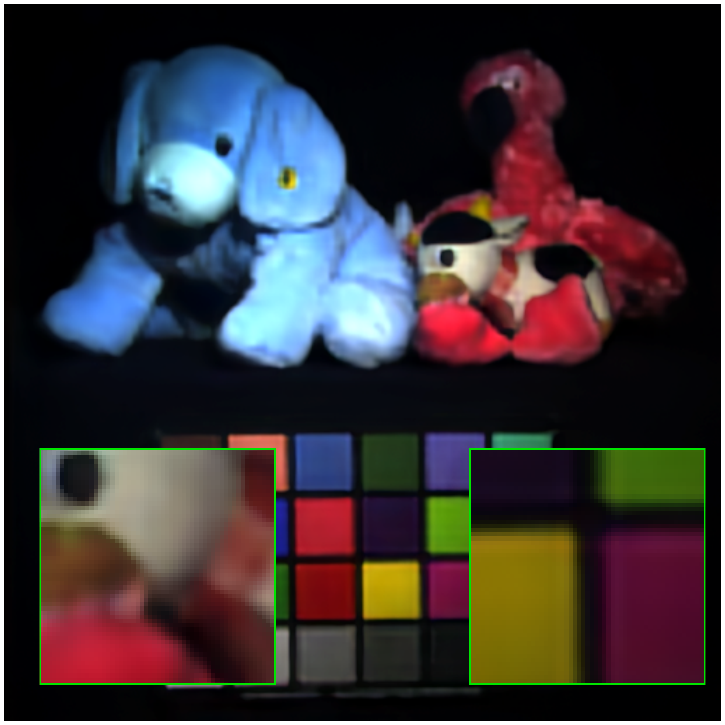}
		\caption{Deep HS Prior~\cite{Sido19}}
	\end{subfigure}\hfill
	\begin{subfigure}{0.19\linewidth}
		\centering
		\includegraphics[width=\linewidth]{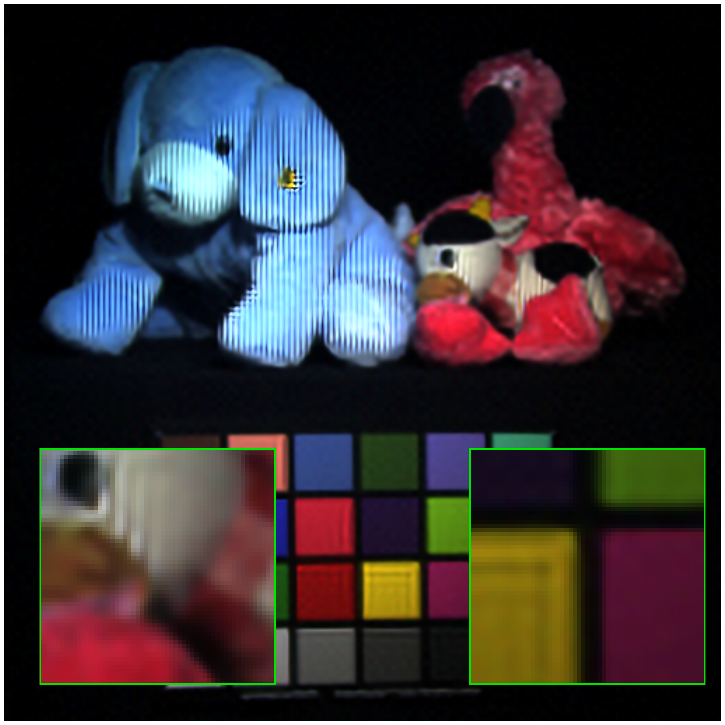}
		\caption{DPHSIR~\cite{Lai22}}
	\end{subfigure}\hfill
	\begin{subfigure}{0.19\linewidth}
		\centering
		\includegraphics[width=\linewidth]{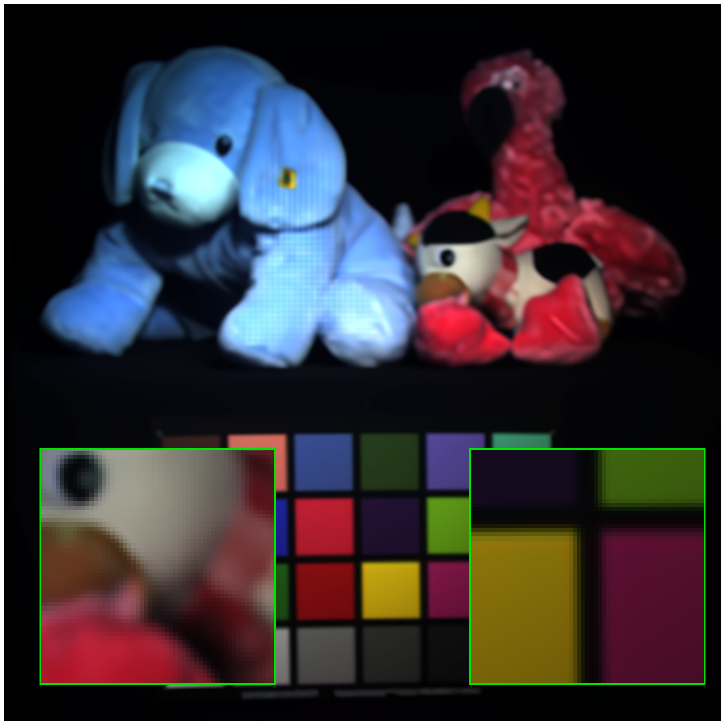}
		\caption{Proposed}
	\end{subfigure}
	
	\begin{subfigure}{0.19\linewidth}
		\centering
		\includegraphics[width=\linewidth]{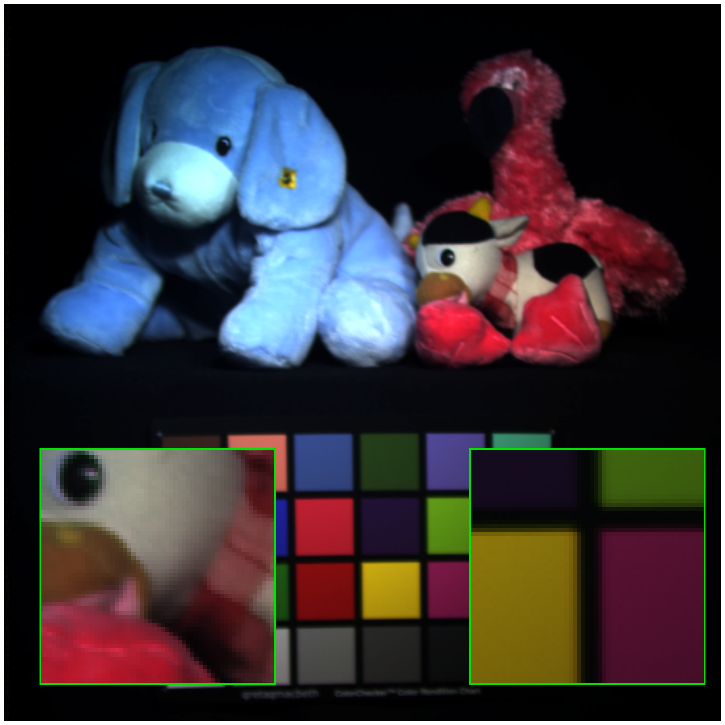}
		\caption{Reference}
	\end{subfigure}\hfill
	\hspace{0.1pt}
    \raisebox{0.6\height}{
        \includegraphics[width=0.025\linewidth]{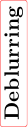}
    }
	\hspace{0.05pt}
	\begin{subfigure}{0.19\linewidth}
		\centering
		\includegraphics[width=\linewidth]{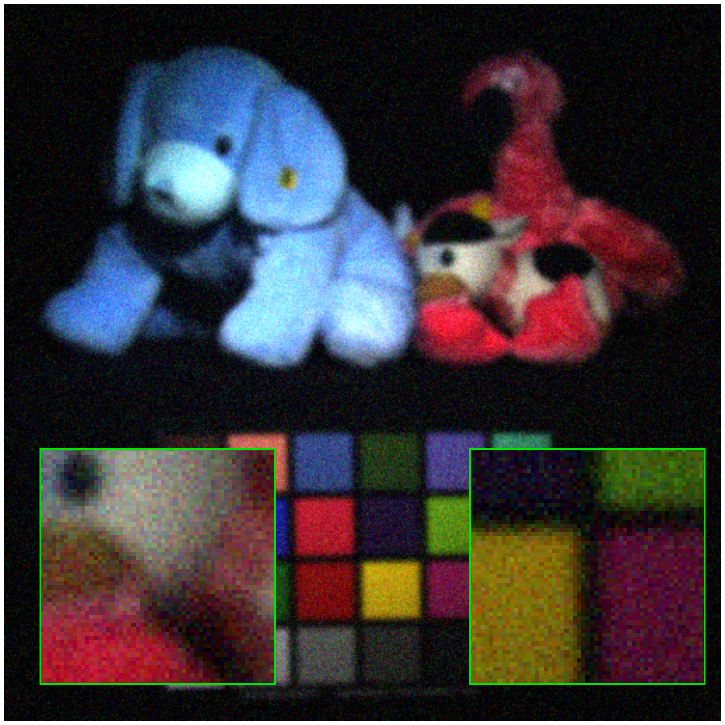}
		\caption{Observation $\mathbf{y}$}
	\end{subfigure}\hfill
	\begin{subfigure}{0.19\linewidth}
		\centering
		\includegraphics[width=\linewidth]{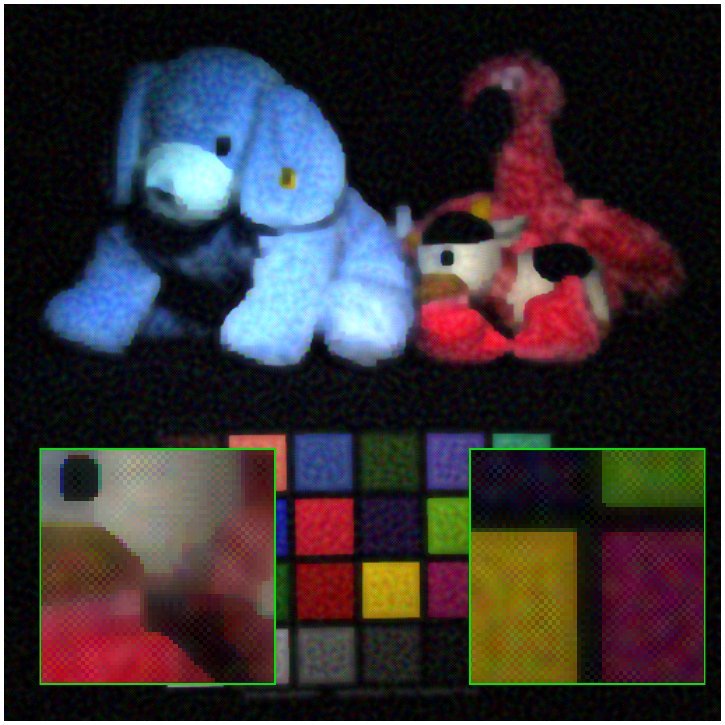}
		\caption{ADMM TV}
	\end{subfigure}\hfill
	\begin{subfigure}{0.19\linewidth}
		\centering
		\includegraphics[width=\linewidth]{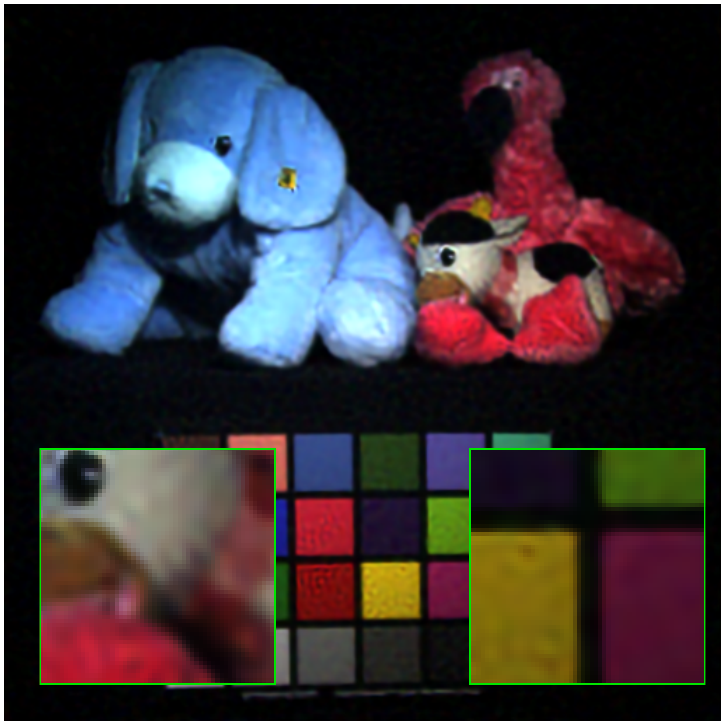}
		\caption{DPHSIR}
	\end{subfigure}\hfill
	\begin{subfigure}{0.19\linewidth}
		\centering
		\includegraphics[width=\linewidth]{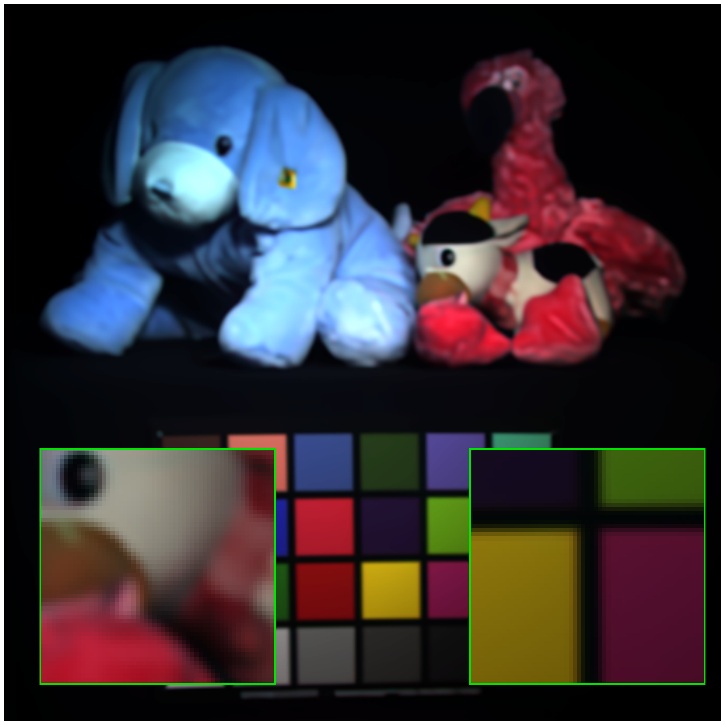}
		\caption{Proposed}
	\end{subfigure}\hfill
	

	\caption{Visual comparison of the proposed reconstruction method with state-of-the-art approaches for an image from the CAVE dataset. Observations and reconstructions are grouped by task, as indicated by the vertical labels on the left, and include zoomed-in regions for detailed visual inspection.}
	\label{fig:methods}
\end{figure*}

\section{Experiments}

\subsection{Experimental setup}

We evaluate our method on the CAVE~\cite{Yasu10} and Harvard~\cite{Chak11} hyperspectral datasets, which consist of natural scenes with 31 spectral bands covering wavelengths from $400\, \mathrm{nm}$ to $700 \mathrm{nm}$ at $10 \mathrm{nm}$ intervals. Images are cropped into non-overlapping $256 \times 256$ patches and split image-wise into 70\% / 15\% / 15\% training, validation, and test sets. To prevent data leakage, all patches originating from the same image are assigned to the same split.

During training, synthetic observations are generated by adding \gls{awgn} with standard deviations $\sigma\in\{0.05,$ $0.10, 0.15, 0.20\}$. These data are used to learn the encoder matrix $\mathbf{E}$, while the pretrained denoiser remains fully frozen. Unless otherwise stated, we use the RGB DRUNet as the low-dimensional denoiser. For the proposed method, we set 
$K=11$, resulting in an encoding matrix $\mathbf{E}\in\mathbb{R}^{33 \times 31}$. At test time, images are degraded according to the task-specific linear observation model (e.g., identity for denoising, downsampling for \gls{sisr}), followed by the addition of \gls{awgn} at the corresponding noise level.
Reconstruction quality is evaluated using \gls{psnr}, \gls{sam}, and \gls{ssim}, which respectively measure numerical accuracy, spectral consistency, and perceptual quality. We additionally report the average processing time per patch.

\subsection{Method comparison}

For the proposed approach, we integrate the hyperspectral denoiser within a \gls{hqs}~\cite{Gema95} framework, using the same parameterization as DPIR~\cite{Zhan22}. For the Harvard dataset, we report results obtained with a simplified, untrained identity encoder, in order to highlight the effectiveness of out-of-the-box configurations.

We consider three representative degradation scenarios. \textit{Denoising} is simulated by adding \gls{awgn} with standard deviations $\sigma=0.10$ and $\sigma=0.20$. \textit{Deblurring} is modeled as convolution with a Gaussian blur kernel followed by \gls{awgn} with $\sigma=0.05$. \textit{\gls{sisr}} is simulated by spatial downsampling by a factor of 4, implemented as Gaussian low-pass filtering followed by uniform subsampling; a small amount of noise ($\sigma=0.005$) is added to avoid trivial inversion.

For denoising, we compare against several state-of-the-art hyperspectral denoisers. For task-agnostic restoration (deblurring and \gls{sisr}), we include representative baselines spanning classical variational methods (ADMM-TV), deep image prior approaches (Deep HS Prior~\cite{Sido19}), \gls{pnp} methods (DPHSIR~\cite{Lai22}), and end-to-end learning-based \gls{sisr} networks (HSISR~\cite{Li22d}).
Among these, DPHSIR is the closest in spirit to our approach, as it also relies on a \gls{pnp} formulation, but employs a hyperspectral-specific denoiser (GRUNet~\cite{Roy24}). We therefore also report GRUNet's standalone denoising performance for completeness. Quantitative results are reported in \tablename~\ref{tab:denoising} and \tablename~\ref{tab:methods} for denoising and for the remaining tasks, respectively, while representative visual comparisons are shown in \figurename~\ref{fig:methods}.

The proposed method achieves strong denoising performance, with especially consistent gains in \gls{sam}, indicating improved spectral fidelity. Visual results confirm that spatial structures are preserved even at high noise levels. On CAVE, this is achieved without preprocessing beyond normalization to $(0,1)$, suggesting that the gains are not merely due to dataset-specific adaptation. 
%
For deblurring and \gls{sisr}, the gains are even more pronounced. This likely stems from DRUNet being explicitly designed for \gls{pnp} restoration, allowing robust artifact suppression throughout the \gls{hqs} iterations. This observation especially holds true when compared to competitive task-agnostic methods (Deep HS Prior, DPHSIR), which operate under our same assumptions.


%
Overall, these results demonstrate that reusing pretrained RGB denoisers with structured spectral projections provides a competitive and versatile alternative to hyperspectral-specific models across a range of restoration tasks.

\begin{table*}[t]
	\centering
	\caption{Denoising performance on the CAVE and Harvard datasets for the denoising case with standard deviations $\sigma=0.10$ and $\sigma=0.20$.
		Results are reported as mean $\pm$ standard deviation. Best results in bold.}
        \resizebox{\textwidth}{!}{%
	\begin{tblr}{
			colspec = {c l c c c c c c c},
			row{1,2} = {font=\bfseries},
			column{3-10} = {mode=math},
			hline{2,3,10},
			vline{2,3,6,9},
			rowsep=0.1pt
		}
		\SetCell[r=2]{l}{} 
        & \SetCell[r=2]{l}{Method} 
        & \SetCell[c=3]{c} \mathrm{CAVE} & &
        & \SetCell[c=3]{c} \mathrm{Harvard} & &
        & 
        \\
		&
		& \mathrm{PSNR}\;\uparrow
		& \mathrm{SAM}\;\downarrow
		& \mathrm{SSIM}\,\uparrow
		& \mathrm{PSNR}\;\uparrow
		& \mathrm{SAM}\;\downarrow
		& \mathrm{SSIM}\,\uparrow
		& \mathrm{Time\,(s)}
		\\
	
		\SetCell[r=7]{c}{\rotatebox{90}{$\sigma = 0.10$}}
		
		
		& SST~\cite{Li23c}
		& 36.35 \pm 2.10 & 11.59 \pm  2.90 & 0.9607 
		& 28.63 \pm 5.57 & 10.90 \pm  3.83 & 0.7527 
		& 0.7712
		\\
		
		& SERT~\cite{Li23d}
		& 38.04 \pm 2.07 &  8.73 \pm  2.50 & 0.9685 
		& 28.74 \pm 5.57 & 10.38 \pm  3.86 & 0.7520 
		& \boldsymbol{0.0655}
		\\
		
		& MAC-Net~\cite{Xion22}
		& 33.20 \pm 0.65 & 26.37 \pm  8.17 & 0.8251 
		& 27.93 \pm 4.43 & 13.78 \pm  5.78 & 0.7251 
		& 2.936
		\\
		
		& T3SC~\cite{Bodr21}
		& 34.16 \pm 2.94 & 14.93 \pm  3.99 & 0.8860 
		& 28.07 \pm 5.05 & 11.82 \pm  3.33 & 0.7279 
		& 0.1850
		\\
		
		& GRUNet~\cite{Roy24}
		& 39.47 \pm 1.47 &  9.28 \pm  3.04 & 0.9636 
		& 29.39 \pm 5.55 & 10.09 \pm  3.72 & 0.7627 
		& 0.1942
		\\
		
		& Deep HS Prior~\cite{Sido19}
		& 29.05 \pm 2.40 & 13.28 \pm  4.85 & 0.8576 
		& 24.46 \pm 4.46 & 12.67 \pm  4.97 & 0.5554 
		& 37.80
		\\
		
		
		
		
		& Proposed
		& \boldsymbol{44.61 \pm 1.88} &  \boldsymbol{4.89 \pm 1.95} & \boldsymbol{0.9805} 
		& \boldsymbol{30.51 \pm 4.26} &  \boldsymbol{8.52 \pm 2.65} & \boldsymbol{0.8219} 
		& 0.2817
		\\	

		\SetCell[r=7]{c}{\rotatebox{90}{$\sigma = 0.20$}}
		
		
		& SST
		& 34.82 \pm 1.65 & 14.53 \pm 3.71 & 0.9438 
		& 27.70 \pm 5.12 & 11.44 \pm 3.83 & 0.7298 
		& 0.7637
		\\
		
		& SERT
		& 36.39 \pm 1.67 & 11.49 \pm 3.48 & 0.9530 
		& 27.83 \pm 5.13 & 10.98 \pm 3.84 & 0.7315 
		& \boldsymbol{0.0671}
		\\
		
		& MAC-Net
		& 27.77 \pm 0.23 & 44.39 \pm 13.57 & 0.5911 
		& 25.11 \pm 3.16 & 18.75 \pm 8.82 & 0.6333 
		& 3.708
		\\
		
		& T3SC
		& 33.66 \pm 2.62 & 15.71 \pm 3.94 & 0.8714 
		& 27.49 \pm 4.77 & 12.10 \pm 3.35 & 0.7077 
		& 0.1826
		\\
		
		& GRUNet
		& 37.17 \pm 1.20 & 13.45 \pm 4.52 & 0.9395 
		& 28.27 \pm 5.14 & 10.78 \pm 3.79 & 0.7317 
		& 0.1971
		\\
		
		& Deep HS Prior
		& 28.48 \pm 1.34 & 16.16 \pm 3.71 & 0.8302 
		& 25.14 \pm 4.72 & 11.07 \pm 3.56 & 0.6069 
		& 37.79
		\\
		
		
		
		
		& Proposed
		& \boldsymbol{41.91 \pm 1.82} &  \boldsymbol{6.32 \pm 2.30} & \boldsymbol{0.9694} 
		& \boldsymbol{28.45 \pm 4.63} &  \boldsymbol{9.99 \pm 3.16} & \boldsymbol{0.7456} 
		& 0.2846
		\\
		
	\end{tblr}%
}
	\label{tab:denoising}
\end{table*}
\begin{table}[t]
	\centering
	\caption{Comparison on CAVE across tasks. Results are reported as mean $\pm$ standard deviation. Best results in bold.}
        \resizebox{\linewidth}{!}{%
	\begin{tblr}{
			colspec = {c l c c c c },
			row{1} = {font=\bfseries},
			column{3-6} = {mode=math},
			vline{2-6},
			hline{2,5},
			rowsep=0.1pt,
			colsep=2pt
		}
		
		& Method
		& \mathrm{PSNR}\;\uparrow
		& \mathrm{SAM}\;\downarrow
		& \mathrm{SSIM}
		& \mathrm{Time} 
		\\
		\SetCell[r=3]{c}{\rotatebox{90}{Deblur}}
		& ADMM TV
		& 27.25\!\pm\!0.28
		& 39.45\!\pm\!11.06
		& 0.3680
		& \boldsymbol{2.297}
		\\
		& DPHSIR~\cite{Lai22}
		& 38.40\!\pm\!1.73
		& 8.18\!\pm\!2.70
		& 0.9518
		& 16.52
		\\
		& Proposed
		& \boldsymbol{39.48\!\pm\!1.99}
		& \boldsymbol{4.44\!\pm\!1.82}
		& \boldsymbol{0.9704}
		& 6.991
		\\
		\SetCell[r=5]{c}{\rotatebox{90}{Superres. $\times 4$}}	
		& ADMM TV
		& 25.42\!\pm\!0.36
		& 41.27\!\pm\!11.24
		& 0.2820
		& 5.743
		\\
		
		& DPHSIR
		& 31.58\!\pm\!4.05
		& 9.04\!\pm\!2.46
		& 0.9210
		& 16.83
		\\
		& Deep HS Pr.
		& \boldsymbol{36.29\!\pm\!2.74}
		& 5.90\!\pm\!0.76
		& 0.9495
		& 676.4
		\\
		& HSISR~\cite{Li22d}
		& 30.58\!\pm\!2.01
		& 10.15\!\pm\!3.36
		& 0.8990
		& \boldsymbol{0.025}
		\\
		& Proposed
		& 32.02\!\pm\!5.90
		& \boldsymbol{4.97\!\pm\!1.43}
		& \boldsymbol{0.9542}
		& 21.34
	\end{tblr}%
}
\label{tab:methods}
\end{table}

\subsection{Ablation studies}

We evaluate several ablation settings corresponding to different choices of the encoder matrix $\mathbf{E}$, while keeping the inner denoiser and decoding strategy fixed.
(1) Sequential injection:
corresponds to $\mathbf{E}=\mathbf{I}_C$, i.e., no spectral projection, where bands are passed sequentially to either a monochromatic or an RGB DRUNet, denoted as ``Seq. Mono'' ($c=1$) and ``Seq. RGB'' ($c=3$), respectively.
(2) Random injection:
uses a permuted identity as an encoder, resulting in random grouping of the bands, passed to the RGB DRUNet, denoted as ``Random''.
(3) PCA projection:
defines $\mathbf{E}$ using spectral principal components of the hyperspectral data, denoted as ``PCA proj.''.
(4) Proposed approach:
learns $\mathbf{E}$ under orthogonality constraints via a QR-based parametrization, evaluated for both $K=1$ and $K=11$, with $c=3$, denoted as ``Grp. $K=1$'' and ``Proposed'', respectively.
The results for the CAVE dataset are shown in \tablename~\ref{tab:ablation}. 

The analysis shows that the RGB denoiser consistently outperforms the monochromatic denoiser, indicating that encoding relationships between adjacent spectral bands is beneficial, even when no explicit spectral mixing is introduced. This observation is further supported by the comparison between sequential and random injection: sequential grouping yields better performance than random ordering, confirming that preserving local spectral continuity is advantageous for the denoising task.
The learned projection further improves all metrics by exploiting spectral correlations through data-driven mixing. However, identity-based configurations remain attractive because they require no hyperspectral adaptation, while \gls{pca} projections perform poorly, suggesting that the projected domain must remain compatible with the RGB denoiser.

\begin{table}[t]
	\centering
	\caption{Comparison of designs for the adapted RGB denoiser on the CAVE dataset for $\sigma=0.10$.
		Results are reported as mean $\pm$ standard deviation. Best results in bold.}
        \resizebox{\linewidth}{!}{%
	\begin{tblr}{
			colspec = {l c c c c},
			row{1} = {font=\bfseries},
			column{2-5} = {mode=math},
			hline{2},
			vline{2},
            rowsep=0.1pt,
			colsep=2pt
		}
		Method
		& \mathrm{PSNR}\;\uparrow
		& \mathrm{SAM}\;\downarrow
		& \mathrm{SSIM}\;\uparrow
		& \mathrm{Time\;(s)}
		\\
		Seq. Mono
		& 41.43\!\pm\!2.06
		& 6.01\!\pm\!2.18
		& 0.9682 
		& 0.7625
		\\
		Seq. RGB
		& 43.33\!\pm\!2.00
		& 5.22\!\pm\!1.98
		& 0.9762 
		& 0.2804
		\\
		Random
		& 42.59\!\pm\!1.92
		& 5.57\!\pm\!2.06
		& 0.9744 
		& 0.2839
		\\
		PCA proj.
		& 33.86\!\pm\!6.60
		& 25.01\!\pm\!5.51
		& 0.6839 
		& 1.717
		\\
		Grp. $K=1$
		& 31.75\!\pm\!2.15
		& 22.85\!\pm\!6.63
		& 0.9206 
		& \boldsymbol{0.0335}
		\\
		Proposed
		& \boldsymbol{44.61\!\pm\!1.88}
		& \boldsymbol{4.89\!\pm\!1.95}
		& \boldsymbol{0.9806} 
		& 0.2817
		\\

	\end{tblr}%
    }
	\label{tab:ablation}
\end{table}

\section{Conclusions}

In this work, we investigated the adaptation of pretrained monochromatic and RGB denoisers for  \gls{hir}. We showed that such denoisers implicitly encode exploitable spectral priors, and we introduced a principled, stable, and task-agnostic framework to lift them to hyperspectral imaging through constrained linear projections. To the best of our knowledge, this is the first projection-based approach that reuses pretrained RGB denoisers with guaranteed stability and seamless compatibility with \gls{pnp} optimization across multiple restoration tasks.
Despite the rich spectral structure of hyperspectral data, priors learned from large-scale RGB datasets remain highly competitive when properly exploited, even without hyperspectral-specific training.
Additionally, treating the RGB denoiser as a frozen black box likely underutilizes its full potential.
%
%
Future work may explore lightweight adaptation strategies to better match hyperspectral statistics and noise models while retaining generality.

\vfill\pagebreak


\bibliographystyle{IEEEbib}
\bibliography{biblio_short.bib}

\end{document}